\documentclass[10pt,twocolumn,letterpaper]{article}

\usepackage{cvpr}
\usepackage{times}
\usepackage{epsfig}
\usepackage{graphicx}
\usepackage{amsmath}
\usepackage{amssymb}
\usepackage{multirow}
\usepackage{subfig}
\usepackage{amsthm}
\usepackage{bbm}
\usepackage[ruled,vlined]{algorithm2e}
\usepackage{makecell}
\usepackage{rotating}
\usepackage{booktabs}
\usepackage{float}

\newcommand{\scal}[1]{\mathit{#1}}
\newcommand{\vect}[1]{\mathbf{#1}}
\newcommand{\matr}[1]{\mathbf{#1}}

\newcommand{\ra}[1]{\renewcommand{\arraystretch}{#1}}

\usepackage[pagebackref=true,breaklinks=true,letterpaper=true,colorlinks,bookmarks=false]{hyperref}

 \cvprfinalcopy 

\def\cvprPaperID{4172} 

\ifcvprfinal\fi
\begin{document}

\title{Motion Segmentation by Exploiting Complementary Geometric Models}

\author{Xun Xu\\
National University of Singapore\\
{\tt\small elexuxu@nus.edu.sg}
\and
Loong Fah Cheong\\
National University of Singapore\\
{\tt\small eleclf@nus.edu.sg}
\and
Zhuwen Li\\
Intel Labs\\
{\tt\small li.zhuwen@intel.com}
}

\maketitle

\begin{abstract}
Many real-world sequences cannot be conveniently categorized as general or degenerate; in such cases, imposing a false dichotomy in using the fundamental matrix or homography model for motion segmentation would lead to difficulty. Even when we are confronted with a general scene-motion, the fundamental matrix approach as a model for motion segmentation still suffers from several defects, which we discuss in this paper. The full potential of the fundamental matrix approach could only be realized if we judiciously harness information from the simpler homography model. From these considerations, we propose a multi-view spectral clustering framework that synergistically combines multiple models together. We show that the performance can be substantially improved in this way. We perform extensive testing on existing motion segmentation datasets, achieving state-of-the-art performance on all of them; we also put forth a more realistic and challenging dataset adapted from the KITTI benchmark, containing real-world effects such as strong perspectives and strong forward translations not seen in the traditional datasets. 

\end{abstract}

\vspace{-0.2cm}
\section{Introduction}
 
Various geometric models have been used in the motion segmentation problem to model the different types of cameras, scenes, and motions. In this problem as commonly set forth, the underlying models are generally regarded as applicable under different scenarios and these scenarios do not overlap. For instance, when the underlying motion is a general motion, fundamental matrix is used to model the epipolar geometry \cite{Jung2014,Li2013}, and when scene-motion is degenerate like a planar scene or a pure rotation, homography is preferred \cite{Dragon2012,Lai2017}. However, the real world scene-motions are in fact not so conveniently divided: they are more typified by near-degenerate scenarios such as a scene that is almost but not quite planar, or a motion that is rotation-dominant but with a non-vanishing translation. In such cases, imposing a false dichotomy in deciding an appropriate model would pose difficulty for subsequent subspace separation. For instance, it is well-known \cite{Goshen2008,sugaya2004geometric,Torr1998} in the case of a scene with dominant-plane, it is easy to find inliers belonging to the degenerate configuration (the plane), but the precision of the resulting fundamental matrix is likely to be very low. Most of the inliers outside the degenerate configuration will be lost, and often the erroneous fundamental matrix will pick up outliers (e.g. from other motion groups). Since this is not a purely planar scene, using homography in a naive manner might fail to group all the inliers together too, resulting in over-segmentation of the subspaces. 

It is also not hard to establish---from a glance of the motion segmentation literature---that of the various models, the fundamental matrix model is generally eschewed, due to the lack of perspective effects in the Hopkins155 benchmark \cite{Tron2007}. However, it is never clearly articulated if the numerical difficulties arising from degeneracies in such approach present insuperable obstacles. And no one has put his/her finger on the exact manner how the resulting affinity matrix is ill-suited for subspace clustering: is it solely due to the degeneracies or are there other factors? Considering that in many real-world applications say, autonomous driving, perspective effects are not uncommon, it surely follows that we should come to a better understanding of the suitability of fundamental matrix (or for that matter, the homography model) as a geometric model for motion segmentation. This, we contend, is far from being the case. For instance, does it follow that if we use the fundamental matrix for wide field-of-view scenes, like those found in the KITTI benchmark \cite{Geiger2013IJRR}, we will get better performance than those using homography? We have in fact as yet no reason to believe that this will be the case, judging by the way how the various algorithms based on affine model still outperform those based on fundamental matrix in individual Hopkins sequences that have larger perspectives (though admittedly still moderate). Indeed, from the results we obtained on the KITTI sequences that we adapted for testing motion segmentation in real-world scenarios, the superiority of the homography-based methods is again observed. Thus, one might naturally ask what factors other than degeneracies are hurting the fundamental matrix approach? And why is the homography matrix approach holding its own in wide perspective scenes, when it possesses none of the geometrical exactness of the fundamental matrix? 

In the remainder of this section, we will briefly investigate the suitability of homography and fundamental matrices ($\mathbf{H}$ and $\mathbf{F}$ respectively) as a geometric model for motion segmentation. We shall henceforth denote the affinity matrices generated by $\mathbf{H}$ and $\mathbf{F}$ as $\mathbf{K_H}$ and $\mathbf{K_F}$ respectively.

\subsection{Success roadmap of $\mathbf{H}$}

The preceding paragraphs have already alluded to the fact that the affinity matrix $\mathbf{K_H}$ may not exhibit high intra-cluster cohesion (due to lack of strong affinity between different planes of the same rigid motion), and thus might lead one to be skeptical of its adequacy for the purpose of motion segmentation. In the Hopkins155 dataset, this is not an overriding concern since most of the sequences have small field-of-view and perhaps the scene is sufficiently far away to be well approximated by a plane; these approximations are seemingly borne out by the good empirical results obtained by a wide variety of approaches based on affine subspace or homography matrix. The recent homography-based method \cite{Lai2017} boasts state-of-the-art performance with a mean error of $0.83\%$. The low error attained is noteworthy given that there are actually some Hopkins sequences with non-negligible perspective effects; we feel that this phenomenon warrants a better explanation than the reasoning offered so far. 

\begin{figure}[!ht]
\begin{center}
\includegraphics[width=0.97\linewidth]{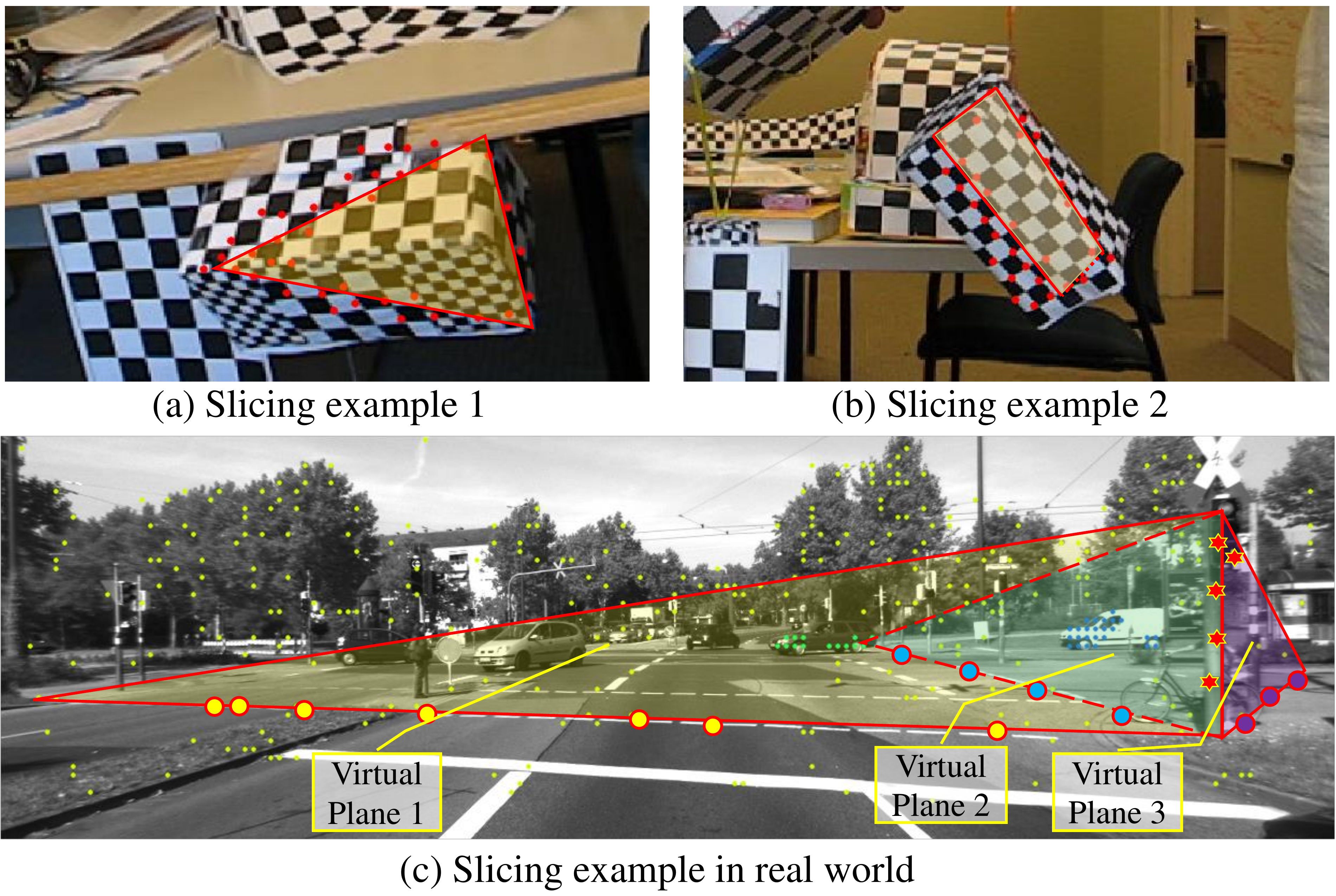}
\caption{Illustration of slicing effect of homography. (a-b) Red dots indicate inlier points of a hyopthesis. All points lie on a virtual plane (a slice of the cube) highlighted in yellow. (c) Virtual planes are highlighted as triangles with points in the same color as inliers. } 
\label{fig:planar_slice}
\vspace{-0.9cm}
\end{center}
\end{figure}

The success can be attributed to the many planar slices induced by the homography hypothesizing process; these are not necessarily actual physical planes in the scenes (see the slices in Fig.~\ref{fig:planar_slice} (a-b)) but as long as these virtual planes belong to the same rigid motion, it is evident that they can be fitted with a homography. Such slicings of the scene create strong connections between points across multiple real planar surfaces and result in a much less over-segmented affinity matrix $\mathbf{K_H}$. If the scene contains only compact objects or piecewise smooth structures, then such connectivity created is sufficient to bind the various surfaces of a rigid motion together. However, in the real world sequences, when the above conditions are not satisfied, we suspect that this may not be adequate. Fig.~\ref{fig:planar_slice}(c) illustrates a background comprising an elongated object (a traffic light) and the marking on the road. It is clear that in this case, while one can form virtual planar slices as before, the resulting connectivity is much lower (most if not all of the slices cannot connect large segments of both these elements simultaneously, unlike those in Fig.~\ref{fig:planar_slice} (a-b)).

\subsection{Problems with $\mathbf{F}$}

Besides the degeneracy issues that are well-known from the classical structure from motion literature, we suggest that another root problem with the fundamental matrix approach for the motion segmentation problem lies precisely in the fact that it is an all-encompassing model that captures all types of scene-motion configurations. The risk of such a complicated model for the subsequent clustering and model-selection task is not difficult to surmise. The richness of characterization renders it likely to capture any correlation between different rigid motions. Therefore it is more likely to cause overlapping the subspaces of different rigid motions than simpler models, e.g. homography. 
However, we find ourselves asking, is it not possible that $\matr{F}$ also offers the greatest scope for forming the best correct view, given that it starts with a geometrically correct model that the homography model can hardly be, and the former must have thus captured much of what is correct? It perhaps just requires some nudge in the correct direction for us to reclaim the performance that ought be had for $\mathbf{K_F}$. From this standpoint, even when we are confronted with a general scene-motion with no degeneracies, there is still an important 
reason for keeping the homography model---to midwife the unborn view of $\mathbf{K_F}$.

\subsection{Proposed solution}
We have been at pains to point out that many real-world sequences cannot be classified into neat categories such as general or degenerate scene-motions and thus cannot be adequately addressed by any single model such as $\mathbf{H}$ or $\mathbf{F}$. We have also discussed the defects of the fundamental matrix approach and conjectured that even though the resulting $\mathbf{K_F}$ may not have committed itself to any definite view on the potential clusters, its full potential could perhaps be realized if we judiciously harness information from a simpler model such as $\mathbf{H}$. From these considerations, we propose a multi-view\footnote{Note that this ``view'' here refers to the view from the standpoint of a model and should not be confused with the camera viewpoint.} spectral clustering framework that synergistically combines these multiple models together. As there is no definite consensus on how best to combine several views together for spectral clustering, we evaluate a few extant fusion schemes. 
By doing so, we make sure that our findings are not an artifact of a particular fusion scheme. As we will show later, the performance of the fundamental matrix approach can be substantially raised using the improved $\mathbf{K_F}$. 
We hasten to add that one should not over-claim the potential gains of this fundamental matrix approach. When the scene contains substantial amount of degeneracies, as real scenes are apt to be, it is always better to rely on the combined view for the best performance. That is, one should seek a common spectral embedding that takes into account both the improved $\mathbf{K_F}$ and the improved $\mathbf{K_H}$.

To summarize, the contributions of our paper are as follows. First, we contribute to an understanding of the strengths and drawbacks of homography and fundamental matrices as a geometric model for motion segmentation. We then propose using affinity matrix fusion as a means of dealing with real-world effects that are often difficult to model with a pure homography or fundamental matrix. 
Finally, we perform extensive testing on existing motion segmentation datasets, achieving state-of-the-art performance on all of them; we also put forth a more realistic and challenging dataset adapted from the KITTI benchmark, containing real-world effects such as strong perspectives and strong forward translations not seen in the traditional datasets.

\section{Related Work}
A long line of works have studied the motion segmentation problem from different perspectives. They can be divided into two major groups: those based on a hypothesis-and-test paradigm and those that are more analytic rather than hypothesis-driven. Into the latter camp falls a wide variety of approaches, including factorization \cite{boult1991factorization,costeira1998multibody, gear1998multibody,gruber2004multibody,tomasi1992shape}, algebraic method \cite{Rao2010,vidal2004motion,vidal2005generalized,Vidal2008}, affinity matrix \cite{lauer2009spectral,yan2006general}, including those constructed from sparse \cite{Elhamifar2013} and low-rank \cite{liu2013robust} representations. They typically assume that the input is made up of a union 
of motions of specific types, with only a few works \cite{goh2007segmenting,Rao2010} that can handle mixed types of motions. These analytic approaches are rightly praised for their elegance but become awkward in dealing with real world signals that are often drawn from mixed multiple manifolds. In contrast, works in the former category, being hypothesis-driven, are naturally more suited to handling mixed models. This is exemplified in the earlier works such as \cite{sugaya2004geometric,Torr1998} which explicitly decide on whether $\mathbf{F}$ or $\mathbf{H}$ is better suited as a motion model in the face of possibly degenerate scene-motion configuration, but these works are applied to cases where the background is by far the most dominant group in the scene. 
Subsequent hypothesis-and-test methods \cite{Chin2009,chin2010accelerated,lazic2009floss} dealing with the realistic Hopkins155 \cite{Tron2007} sequences almost as a rule ignore the more complex fundamental matrix (or equivalently the perspective projection model) altogether, possible reasons being the computational complexity issues posed by the outliers under the fundamental matrix model and/or the lack of perspective effects in the Hopkins sequences. Thus, these later works do not concern themselves with the problem of dealing with mixed types of models. Our approach differs from the above works in that not only do we allow for mixed types of models, we also do not impose a dichotomous decision on what is an appropriate model.  


Spectral clustering has been an attractive tool for clustering data  \cite{von2007tutorial}. Under this framework there are roughly two genres. The first kind discovers an optimal combination to aggregate multiple affinity matrices (kernels) for spectral clustering \cite{Huang2012,lange2006fusion,Wang2013}. However such combination is often non-trivial to discover. Alternatively, studies have been carried out on discovering a consensus on multiple kernels. In particular, the co-regularization scheme \cite{Kumar2011} was proposed to force data from different views to be close to each other in the embedding space for clustering. 
 Few if any of the existing approaches can guarantee superiority to the simple approach---kernel addition. In this work, we start our evaluation with this simplest baseline and then reveal its relation with the co-regularization schemes. {We also evaluate a custom-built version incorporating a subset constraint that preserves the true hierarchical structure of the affinity matrices induced by different geometric models}.

\section{Methodology}

In this section, we first describe the geometric models used for motion segmentation and their hypothesis formation process. We then explain how the affinities between feature points are encapsulated in the ORK kernel\cite{Lai2017}. Finally, we explain the extension from single-view to multi-view clustering. In particular, we elaborate the relation between kernel addition and co-regularization for generic multi-kernel clustering, and we describe how the geometric relation that exists between models can be used to formulate a custom-made subset constrained multi-view clustering.

\subsection{Geometric Model Hypothesis}

Denote the observations of tracked points throughout $\scal{F}$ frames as $\{\vect{x}_{i}\}_{f=1\cdots F}$. We then randomly sample a minimal number of $p$ such points visible in a pair of frames and use them to fit a hypothesis of the model. The models tested include the fundamental matrix $\mathbf{F}$, homography $\mathbf{H}$, as well as the affine matrix $\mathbf{A}$. The reason for including the affine matrix model is because many existing datasets contain sequences with very weak perspective so this simpler model might be numerically more stable. For the three models $\mathbf{F}$, $\mathbf{H}$, and $\mathbf{A}$, the respective values for $p$ are 8, 4, and 3. The parameters of the model are estimated via linear algorithms \cite{hartley2003multiple} and $500\times F$ hypotheses are sampled for each type of geometric model. 


\subsection{Affinity Captured as Ordered Residual Kernel}

Given multiple hypotheses $\{\matr{Y}_k\}_{k=1\cdots K}$ generated from a particular model (affine, homography or fundamental matrix), we first compute for each data point the residual to all these hypotheses $\{d(\vect{x}_i,\matr{Y}_k)\}_{k=1\cdots K}$ in terms of their Sampson errors \cite{hartley2003multiple}. The affinity between two features is captured in the correlation of preference for these hypotheses. Specifically, we can define the correlation in terms of the co-occurrence of points among all hypotheses. That is, if we define the indicator of point $\vect{x}_i$ being the inlier of all hypotheses $\{\matr{Y}_k\}$ as $\vect{o}_i\in\{0,1\}^{\scal{K}}$, then the co-occurrence between two points is written as $k_{ij}=\vect{o}_i^\top\vect{o}_j$. 
However, the threshold $\tau$ needed to determine when a data is an inlier (i.e. $\vect{o}_i=\mathbbm{1}(d(\vect{x}_i,\matr{H}_k)<\tau)$) is not easy to set, due to the potentially disparate range of motions present in different sequences.
The ordered residual kernel (ORK) \cite{Chin2009,Lai2017} was proposed to deal with this issue. Instead of fixing a threshold, the ORK sorts the residual in ascending order $\{\hat{d}_{i1} \quad \hat{d}_{i2} \cdots \hat{d}_{iK}\}$ where $\forall k : \hat{d}_{ik}<=\hat{d}_{ik+1}$. An adaptive threshold is then selected as the top $h$-th residual, i.e. $\tau_i=\hat{d}_{ih}$. The ORK kernel is also known to be resilient to serious sampling imbalance, an important advantage in real-world scenes where background is usually very large. Therefore, we adopt the ORK kernel to encapsulate the affinities between feature points. After constructing the affinity matrix, we normalize the affinities by dividing all $k_{ij}$ entries by the number of frames where both feature points $i$ and $j$ are visible. This step removes the weighting balance caused by incomplete trajectories. 
Finally, as is customary in motion segmentation works, we subject the affinity matrix to a sparsification step; we use the $\epsilon$-neighborhood scheme of \cite{Lai2017} for this purpose.

\subsection{Spectral Clustering for Motion Segmentation}

We are now ready to use spectral clustering to recover the
clusters. We first review the single view spectral clustering
problem and then extend it to multi-view clustering.

\vspace{-0.2cm}
\subsubsection{Single-View Spectral Clustering}


Given the single affinity matrix $\matr{K}$, the normalized Laplacian $\matr{L}=\matr{I}-\matr{D}^{-0.5}\matr{K}\matr{D}^{-0.5}$ is first computed, where $\matr{D}$ is the degree matrix. The following objective is then set up to eigendecompose $\matr{L}$: 

\begin{equation}
\min\limits_{\matr{U}} tr\left(\matr{U}^\top \matr{L} \matr{U}\right), \quad s.t. \matr{U}\matr{U}^\top=\matr{I}
\end{equation}
where $tr\left(\cdot\right)$ is the trace operator. The spectral embedding $\matr{U}\in\mathbbm{R}^{N\times M}$ can be efficiently solved and then treated as a new feature representation of the original points. A separate K-means step is then fed with the first $M$ dimensions of the normalized $\matr{U}$ for grouping points into $M$ motions.

\subsubsection{Multi-View Spectral Clustering}

With multiple views provided by the different types of motion models, we have now at our disposal multiple affinity matrices. We explore two generic and one custom-made multi-view spectral clustering schemes to fuse the multiple sources of information together for clustering.
\vspace{-0.2cm}
\paragraph{Kernel Addition} A naive way to fuse information from heterogeneous sources for clustering is by kernel addition \cite{Kumar2011}. Given affinity matrices induced by heterogeneous sources $\{\matr{K}_v\}_{v=1\cdots V}$, kernel addition yields a fused kernel by summing up each individual kernel $\matr{K}=\sum_v \matr{K}_v$. With the corresponding Laplacian matrices written as $\matr{L}_v=\matr{I}-\matr{D}^{-0.5}_v\matr{K}_v\matr{D}^{-0.5}_v$, the objective for kernel addition can be written as,

\vspace{-0.2cm}
\begin{equation}
  \resizebox{0.7\linewidth}{!}{$
\begin{split}
&\min\limits_{\matr{U}} tr(\vect{U}^\top\sum_v\matr{L}_v\vect{U}),\quad s.t. \vect{U}^\top\vect{U}=\matr{I}\\
\Rightarrow &\min\limits_{\{\matr{U}_v\}} \sum_v tr(\vect{U}_v^\top\matr{L}_v\vect{U}_v),\quad s.t. \vect{U}_v^\top\vect{U}_v=\matr{I}, \\
&\quad \forall v,w\in\{1,\cdots V\}: \vect{U}_v=\vect{U}_w
\end{split}$}
\end{equation}

 We notice the kernel addition strategy is equivalent to discovering a common spectral embedding $\matr{U}$ among all views. This requirement of having a single consensus embedding can be too strong.

 \vspace{-0.2cm}
 \paragraph{Co-Regularization} Instead of demanding a common embedding, another solution is to include an additional regularization term in the objective function to encourage pairwise consensus between any two spectral embeddings $\matr{U}_v$ and $\matr{U}_w$. This has been studied by \cite{Kumar2011} who introduced a co-regularization term $tr\left(\matr{U}_v\matr{U}_v^\top\matr{U}_w\matr{U}_w^\top\right)$. This trace term returns high value if the new kernel matrix in the spectral embedding space $\matr{U}_v\matr{U}_v^\top$ and $\matr{U}_w\matr{U}_w^\top$ are similar to each other and vice versa. Incorporating the co-regularization term, we obtain the following objective:
\vspace{-0.0cm}
\begin{equation}\label{eq:CoRegObj}
  \resizebox{0.9\linewidth}{!}{$
\begin{split}
\min_{\{\matr{U}_v\}} &\sum_v tr(\vect{U}_v^\top\matr{L}_v\vect{U}_v)-\lambda\sum_v\sum_{w,w\neq v}tr(\matr{U}_v\matr{U}_v^\top\matr{U}_w\matr{U}_w^\top),\\
s.t. &\vect{U}_v^\top\vect{U}_v=\matr{I}\\
\end{split}$}
\end{equation}

We can interpret the co-regularization scheme as a relaxed version of kernel addition. By increasing the penalty coefficient $\lambda$, the co-regularization scheme will approach kernel addition as all embeddings are forced to approach each other. This model is termed as pairwise co-regularization by \cite{Kumar2011} as the co-regularization term comprises of all pairs of spectral embeddings. The co-regularization model can be efficiently solved by initializing each view $\matr{U}_v$ separately in the same way as single-view spectral clustering. Then we recursively update each view with all other views fixed. When solving a single view, the problem becomes a standard eigendecomposition problem. 
After convergence, we can concatenate the new spectral embedding of all views to produce an extended feature for the K-means step.

\vspace{-0.2cm}
\subsubsection{Subset Constrained Multi-View Spectral Clustering}

The above two multi-view spectral clustering schemes are generic fusion methods that do not exploit any relation that might exist between the different views. In the specific case of motion segmentation, we know that for any $\matr{H}$ between two views, we can always define a family of $\matr{F}=[\vect{e}]_x\times \matr{H}$ parameterized by a vector $\vect{e}$, where $[\vect{e}]_x$ denotes the skew-symmetric matrix of $\vect{e}$ \cite{hartley2003multiple}. This means a pair of points that are the inliers of a homography should always be the inliers of a certain fundamental matrix. Conversely, if a pair of points are not the inliers of any $\matr{F}$, there is no homography which could take both points as inliers\footnote{We assume in the above two propositions that there are always enough points to fit an $\matr{F}$ if it exists.}. Generally speaking, we should expect that if $\mathbf{K_A}$, $\mathbf{K_H}$, and $\mathbf{K_F}$ are ideal binary affinity matrices, then $\mathbf{K_A} \le \mathbf{K_H} \le \mathbf{K_F}$. We term this hierarchical relationship the subset constraint. Imposing this constraint will help to further denoise or repair the affinity matrices. We cast this problem as a constrained clustering problem (adapted from \cite{Wang2014}):

\begin{equation}
\resizebox{.7\hsize}{!}
{$
\begin{split}
\min_{\{\matr{U}_v\}} &\sum_v tr\left(\matr{U}_v^\top\matr{L}_v\matr{U}_v\right) - \gamma tr(\matr{U}_v^\top\matr{Q}_v\matr{U}_v),\\ 
s.t. &\matr{U}_v^\top\matr{U}_v=\matr{I},\quad \matr{Q}_v\in \{-1,0,1\}^{N\times N}\\
\end{split}
$}
\end{equation}
\noindent
where the matrix $\matr{Q}_v$ provides the subset constraint for the $v$-th view. For $q_{ij}=1$, the constraint encourages a high inner product $\vect{u}_{vi}^\top\vect{u}_{vj}$ where $\vect{u}_{vi}$ indexes the $i$-th column. This means points $i$ and $j$ are encouraged to fall into the same cluster. For $q_{ij}=-1$, the constraint encourages a different cluster assignment between $i$ and $j$, and lastly, for $q_{ij}=0$, there is no constraint. For any single view $v$, the constraints $\matr{Q}_v$ is imposed by other views. For example, solving view $\matr{H}$, the positive constraint $q_{ij}$ is inherited from the result of $\matr{K}$; that is, if there is a link between points $i$ and $j$ from $\matr{K_A}$, then the $(i,j)$ entry of $\matr{K_H}$ is encouraged to be 1. On the other hand, the negative constraints come from $\matr{F}$. One could solve this problem using an alternating minimization scheme, but the subset constraint matrix $\matr{Q}_v$ may flip their values from 1 to -1 and vice versa in each alternating step, posing significant difficulties for convergence.

Therefore, we relax $\matr{Q}_v$ to continuous values. Instead of utilizing the discretized results from other views, we use the affinity reconstructed from the spectral embedding $\matr{\hat{K}}=\matr{U}\matr{U}^\top$ to construct $\matr{Q}_v$ as detailed in Eq~(\ref{eq:Qconstraint}). We assume the three views are placed in the order of affine ($v=1$), homography ($v=2$) and fundamental matrix ($v=3$). The final objective is then written as Eq~(\ref{eq:Qconstraint}).

\vspace{-0.2cm}
\begin{equation}
\resizebox{1\hsize}{!}
{$
\begin{split}
&\min_{\{\matr{U}_v\}} \sum_v tr\left(\matr{U}_v^\top\matr{L}_v\matr{U}_v\right) - \gamma tr(\matr{U}_v^\top\matr{Q}_v\matr{U}_v),\quad s.t. \matr{U}_v^\top\matr{U}_v=\matr{I}, \\
&\matr{Q}_v = \begin{cases}
 \mathbbm{1}\left(\matr{\hat{K}}_{v+1} <0\right)\circ \matr{\hat{K}}_{v+1}, \quad v=1 \\
\mathbbm{1}\left(\matr{\hat{K}}_{v-1} >0\right)\circ\matr{\hat{K}}_{v-1} + \mathbbm{1}\left(\matr{\hat{K}}_{v+1} <0\right)\circ \matr{\hat{K}}_{v+1}, \quad v=2 \\
\mathbbm{1}\left(\matr{\hat{K}}_{v-1} >0\right)\circ \matr{\hat{K}}_{v-1}, \quad v=3
\end{cases}
\end{split}
$}\label{eq:Qconstraint}
\end{equation}
\noindent
where $\circ$ represents element-wise multiplication and $\mathbbm{1}\left(\cdot\right)$ is the indicator function. The subset constraint means for view $\matr{A}$ ($v=1$), only the negative constraint from $\matr{H}$ is applied, for view $\matr{H}$, both positive and negative constraints from $\matr{A}$ and $\matr{F}$ are applied respectively.  The final problem can be solved by optimizing each view $\matr{U}_v$ in an alternating fashion. We summarize the whole procedure in Algorithm~\ref{alg:Subset}.

\IncMargin{1em}
\begin{algorithm}
\SetKwData{Left}{left}\SetKwData{This}{this}\SetKwData{Up}{up}
\SetKwFunction{Concatenate}{Concatenate}\SetKwFunction{Kmeans}{K-means}
\SetKwInOut{Input}{input}\SetKwInOut{Output}{output}
\Input{\small{Kernel matrices $\{\matr{K}_v\}$, no. of motion $M$ and $\gamma$}}
\Output{Rigid motion index $s$}
\emph{Initialize Spectral Embedding}\\
\For{$v \leftarrow 1$ \KwTo $V$}
{
	Compute Laplacian matrix $\matr{L}_v=\matr{I}-\matr{D}^{-0.5}_v\matr{K}_v\matr{D}^{-0.5}_v$\;
	$\matr{U}_v \leftarrow$ first $M$ eigenvectors of $\matr{L}_v$\;
}
\emph{Subset Constrained Spectral Clustering}\\
\While{Not Converged}
{
\For{$v \leftarrow 1$ \KwTo $V$}
{
	Compute $\matr{Q}_v$ following Eq~(\ref{eq:Qconstraint})\;
	Compute constrained Laplacian matrix $\tilde{\matr{L}}_v=\matr{L}_v - \gamma\matr{Q}_v$\;
	$\matr{U}_v \leftarrow$ first $M$ eigenvectors of $\tilde{\matr{L}}_v$\;
}}
\emph{K-means to return index}\\
{$\matr{U}\leftarrow$ \Concatenate{$\matr{U}_1,\cdots\matr{U}_V$} 
}\;
$s\leftarrow$ \Kmeans{$\matr{U}$,$M$}
\caption{\small{Subset Constrained Clustering}\label{alg:Subset}}
\end{algorithm}
\DecMargin{1em}

\subsubsection{Convergence Analysis}

For both co-regularization and subset constrained clustering, we note the objective is not guaranteed to be convex w.r.t. all views' embeddings. Nevertheless, we prove that the co-regularization model guarantees to converge to at least a local minimal. As we solve the problem in an alternating fashion, each step involves solving Eq~(\ref{eq:CoRegObj}) for $v$-th view with all other views fixed, i.e. $\min_{\matr{U}_v}tr\left(\matr{U}_v^\top\left(\matr{L}_v-\lambda\sum_{w,w\neq v}\matr{U}_w\matr{U}_w^\top\right)\matr{U}_v\right)$. Such problem can be efficiently solved by eigen decomposition regardless of the convexity of $\left(\matr{L}_v-\lambda\sum_{w,w\neq v}\matr{U}_w\matr{U}_w^\top\right)$. Therefore, solving all views iteratively results in monotonic decreasing cost until converging to a local minimal. The convergence for subset constrained clustering is, however, not guaranteed due to the constraint matrix $\matr{Q}_v$ changes at each iteration. Nevertheless, experiment results suggest a proper selection of $\lambda$, less than $1e-2$ renders the problem easy to converge.



\vspace{-0.2cm}
\section{Experiment}

We carry out experiments on three extant motion segmentation benchmarks including the Hopkins155 \cite{Tron2007}, the Hopkins12 \cite{Rao2010} for testing incomplete trajectories and MTPV62 \cite{Li2013} for testing stronger perspective effects. For all three datasets, we evaluate the performance in terms of classification error \cite{Tron2007}. We also put forth a new dataset that is adapted from the KITTI benchmark \cite{Geiger2013IJRR}, containing real-world effects such as strong perspectives and strong forward translations not seen in the traditional datasets.

\subsection{Motion Segmentation on Existing Benchmarks}

In this section, we extensively compare single-view and multi-view approaches on Hopkins155 benchmark \cite{Tron2007}. Specifically, for single-view, we evaluate using affine, homography and fundamental matrix as the single geometric model. For multi-view motion segmentation, we evaluated Kernel Addition (KerAdd), Co-Regularization (CoReg) \cite{Kumar2011} and {Subset Constrained Clustering (Subset)}. We fix the regularization parameter $\lambda$ and $\gamma$ at $10^{-2}$. 
We also extensively compare with state-of-the-art approaches, including:  ALC \cite{Rao2010}, GPCA \cite{Vidal2008}, LSA \cite{Yan2006}, SSC \cite{Elhamifar2013}, TPV \cite{Li2013}, T-Linkage \cite{Magri2014}, S$^3$C \cite{Li2015}, RSIM \cite{Ji2016} and MSSC\cite{Lai2017}. The results are presented in Table~\ref{tab:AllPerf}. For those algorithms which do not explicitly handle missing data, we recover the data matrix using Chen's matrix completion approach \cite{Chen2008}.

We make the following observations from the results. Firstly, with regards to the use of homography matrix as a single geometric model, our finding echoes the excellent results of earlier work such as MSSC \cite{Lai2017}. In fact, the simpler affine model has an even lower error figures. Clearly, the stitching argument (via virtual slices) put forth in Section 1 for explaining the success of homography applies to the affine case too, in particular under weak perspective views. For the fundamental matrix as a model, the performance is slightly worse-off. The reasons are manifold: strong camera rotation, limited depth relief, and not least the subspace overlap between different rigid motions, to which this richer fundamental matrix model is particularly susceptible. 
Secondly, after fusing multiple kernels, we saw a boost in performance compared to single-view approaches, e.g. $0.36\%$ error for kernel addition and $0.31\%$ for subset constrained clustering on Hopkins155. Consistent boost in performance can be observed on Hopkins12 and MTPV62 as well. Usually, the fusion can produce the best of all performance regardless of the fusion scheme used. Even the simple kernel addition yields very competitive performance. This provides a strong option for real applications where parameter tuning is not desirable. 

\begin{table*}[htbp]
\ra{0.88}
  \centering
  \caption{Motion segmentation results on Hopkins155, Hopkins12, MTPV62 and KT3DMoSeg datasets evaluated as classification error ($\%$). $^*$The best performing model (RPCA+ALC$_5$ is reported for ALC \cite{Rao2010}). $^{**}$ State-of-the-Art models' performances are reported for the sequences with correct number of motion. `$-$' cells indicate not reported or no public code is available.}\vspace{-0.2cm}
  \resizebox{0.9\textwidth}{!}{
    \begin{tabular}{llllllllllll}
    \toprule
    Models & \multicolumn{3}{c}{Hopkins155 \cite{Tron2007}} & \multicolumn{2}{c}{Hopkins12 \cite{Rao2010}} & \multicolumn{4}{c}{MTPV62 \cite{Li2013}$^{**}$} & \multicolumn{2}{c}{KT3DMoSeg} \\
    \cmidrule(lr){1-1}
    \cmidrule(lr){2-4}
\cmidrule(lr){5-6}
\cmidrule(lr){7-10} \cmidrule(lr){11-12}
    \multicolumn{1}{l}{\textit{\small{State-of-the-Art}}} & \multicolumn{1}{c}{\footnotesize{2 Motion}} & \multicolumn{1}{c}{\footnotesize{3 Motion}} & \multicolumn{1}{c}{\footnotesize{All}} & \multicolumn{1}{c}{\footnotesize{Average}} & \multicolumn{1}{c}{\footnotesize{Median}} & \multicolumn{1}{c}{\thead{Perspective\\ 9 clips}} & \multicolumn{1}{c}{\thead{Missing Data\\ 12 clips}} & \multicolumn{1}{c}{\thead{Hopkins\\ 50 clips}} & \multicolumn{1}{c}{\thead{All\\ 62 clips}} & \multicolumn{1}{c}{\footnotesize{Average}} & \multicolumn{1}{c}{\footnotesize{Median}}\\
    \midrule
    LSA \cite{Yan2006} & 4.23  & 7.02  & 4.86  & -     & -     & -     & -     & -     & - & 38.30 & 38.58 \\
    GPCA \cite{Vidal2008} & 4.59  & 28.66 & 10.02 & -     & -     & 40.83 & 28.77 & 16.20 & 16.58 & 34.60 & 33.95\\
    {ALC \cite{Rao2010}} & 2.40  & 6.69  & 3.56  & 0.89$^*$ & 0.44$^*$ & 0.35  & 0.43  & 18.28 & 14.88 & 24.31 & 19.04\\
    SSC \cite{Elhamifar2013} & 1.52  & 4.40  & 2.18  & -     & -     & 9.68  & 17.22 & 2.01  & 5.17 & 33.88 & 33.54\\
    TPV \cite{Li2013} & 1.57  & 4.98  & 2.34  & -     & -     & 0.46  & 0.91  & 2.78  & 2.37 & - & -\\
    LRR \cite{liu2013robust} & 1.33  & 4.98  & 1.59  & -     & -     & -  & -  & -  & - & 33.67 & 36.01\\
    \multicolumn{1}{l}{T-Linkage \cite{Magri2014}} & 0.86  & 5.78  & 1.97  & -     & -     & -     & -     & -     & - & - & - \\
    S$^3$C  \cite{Li2015} & 1.94  & 4.92  & 2.61  & -     & -     & -     & -     & -     & - & - & - \\
    RSIM \cite{Ji2016} & 0.78 & 1.77 & 1.01 & 0.68 & 0.70 & - & - & - & - & - & - \\
    \multicolumn{1}{l}{MSSC \cite{Lai2017}} & 0.54  & 1.84  & 0.83  & -     & -     & -     & 0.65  & \textbf{0.65}  & \textbf{0.65} & - & - \\
    \midrule
    \textit{Single-View} & \multicolumn{11}{l}{} \\
    \midrule
    Affine & 0.40  & 1.26  & 0.59  & 0.15  & {0.10} &   0.25   & 0.35  & 0.93  & 0.82 & 15.76 & 11.52 \\
    Homography & 0.45  & 1.61  & 0.71  & 0.18  & {0.10} &   0.70   & 0.48  & 1.23  & 1.08 & 11.45 & 7.14 \\
    Fundamental & 1.22  & 7.60  & 1.79  & 1.10  & {0.10} &   5.09  & 2.53  & 4.31  & 3.97 & 13.92 & 5.09 \\
    \midrule
    \textit{\small{Multi-View}} & \multicolumn{11}{l}{} \\
    \midrule
    KerAdd & {0.27} & {0.66}  & 0.36  & {0.11} & \textbf{0.00} &  1.54   & 1.41  & 0.76  & 0.88 & 8.31 & 1.02 \\
    CoReg & 0.37  & {0.75} & {0.46} &   \textbf{0.06}  &  \textbf{0.00}     &   {0.22}   & \textbf{0.30} & {0.83} & {0.73} & \textbf{7.92} & {0.75} \\
     Subset & \textbf{0.23}  & \textbf{0.58} & \textbf{0.31} &   \textbf{0.06}  &  \textbf{0.00}     &   \textbf{0.20}   & \textbf{0.30} & {0.77} & \textbf{0.65} & {8.08} & \textbf{0.71} \\
    \bottomrule
    \end{tabular}%
    }\vspace{-0.2cm}
  \label{tab:AllPerf}%
\end{table*}%

\subsection{Motion Segmentation on KITTI Benchmark}

The limitations of the Hopkins155 dataset are well-known: limited depth reliefs, dominant camera rotations, among others. 
Such a dataset cannot meet the requirements of a benchmark for investigating motion segmentation capability in-the-wild, in particular self-driving scenario where the camera platform is often performing large translation and the scene is considerably more complex. For this reason, we propose a new motion segmentation benchmark based on the KITTI dataset \cite{Geiger2013IJRR}, the KITTI 3D Motion Segmentation Benchmark (KT3DMoSeg). We choose short video clips from the raw sequences of KITTI governed by three principles. Firstly, we wish to study sequences with more significant camera translation so camera mounted on moving cars are preferred. Secondly, we wish to investigate the impact of complex background structure, therefore, scene with strong perspective and rich clutter (in the structure sense) is selected. 
Lastly, we are interested in the interplay of multiple motions, so clips with more than 3 motions are also chosen, as long as these moving objects contain enough features for forming motion hypotheses. 22 short clips, each with 10-20 frames, are chosen for evaluation. We further extract dense trajectories from each sequence using \cite{Sundaram2010} and prune out trajectories shorter than 5 frames. Illustration of sample frames with labelled ground-truth and further details about the dataset (such as the preprocessing of trajectories) are given in the supplementary material.

\begin{center}

\begin{figure*}[t]
\subfloat[Individual error]{\includegraphics[width=0.64\linewidth]{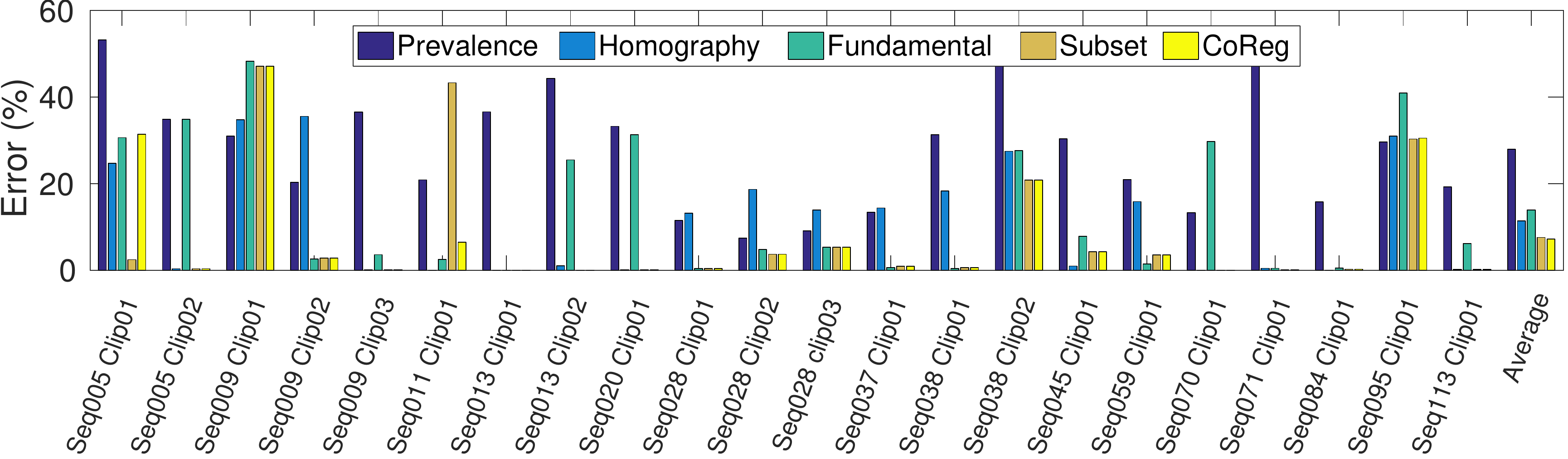}}
\subfloat[CoRegularization]{\includegraphics[width=0.175\linewidth]{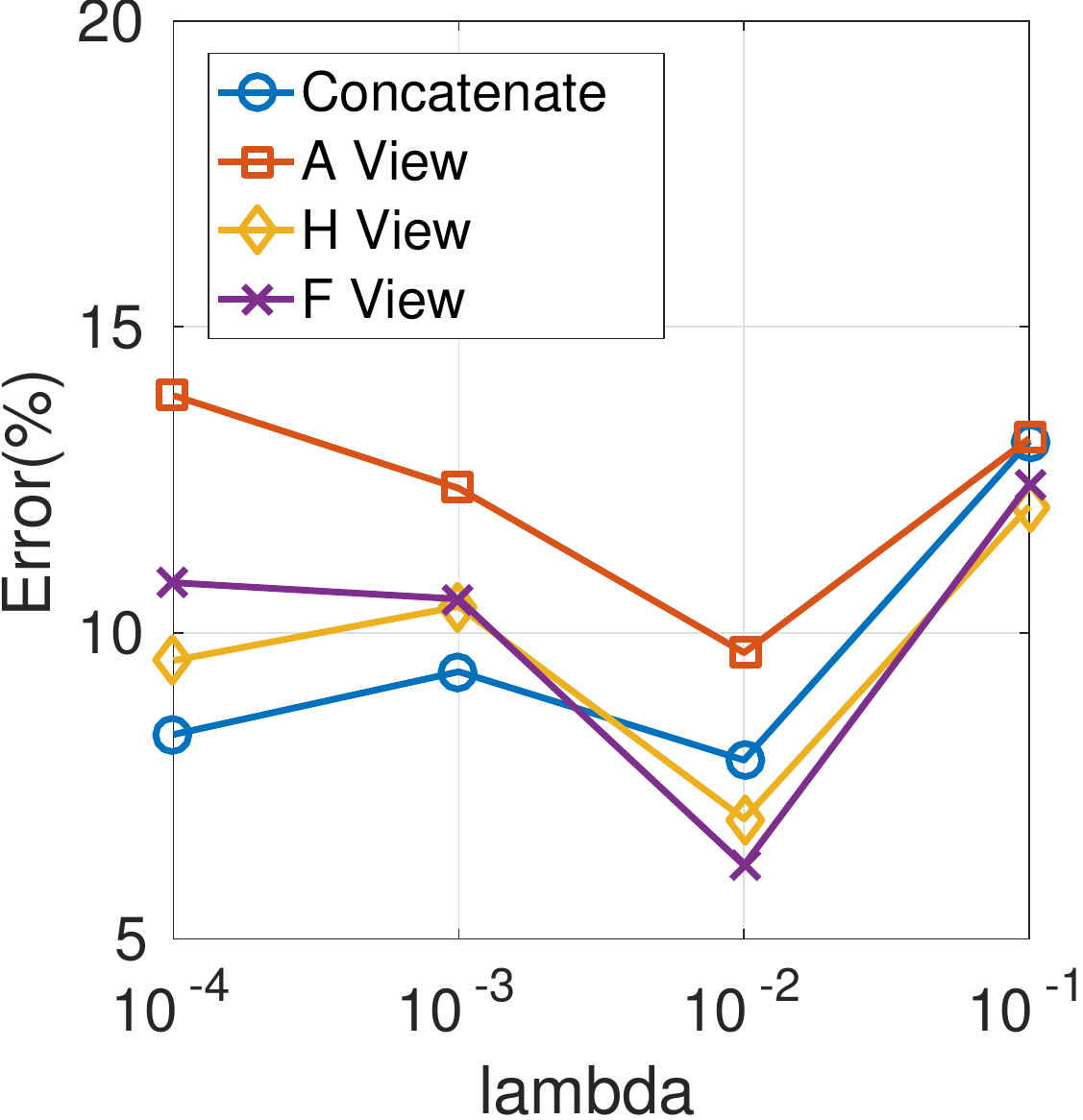}}
\subfloat[SubsetConstrained]{\includegraphics[width=0.175\linewidth]{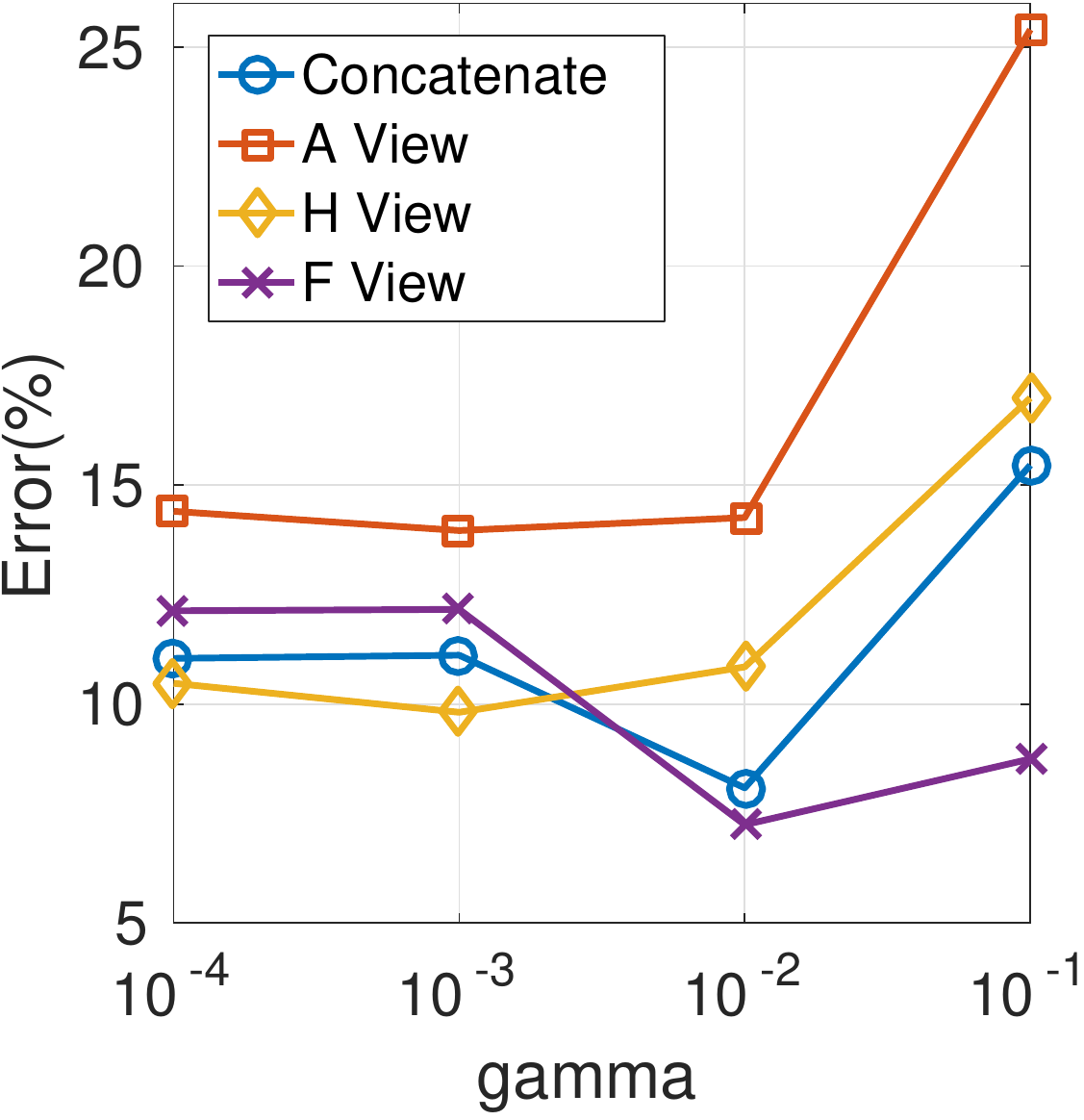}}
\vspace{-0.2cm}
\caption{Classification error on individual sequence and sensitivity to parameters for KT3DMoSeg benchmark.}\label{fig:KTMoSeg_PerfPerSeq}\vspace{-0.2cm}

\end{figure*}
\end{center}

We fit hypotheses on all valid tracking points, i.e. dense background and the evaluation is carried out on subsampled background as introduced in the supplementary. The same set of evaluation as in the preceding subsection is carried out and the results are presented in Tab.~\ref{tab:AllPerf}.  Both average and median classification errors are reported. The performances of the multi-view approaches are again consistently better than those of the single geometric model. Further evaluation on individual sequence is presented in Fig.~\ref{fig:KTMoSeg_PerfPerSeq} (a). To give some context to the performance figures, we use the ``Prevalence'' column to indicate the baseline solution of just assigning every feature as belonging to the prevalent group---the background. The overall performance of this baseline approach is $27.95\%$ which is pretty strong compared to many existing approaches. 
For the more recent and hypothesis-driven approach like MSSC, although we do not have the codes for evaluation, we can get an idea of its performance in KT3DMoSeg by looking at the result of our single-view homography model, due to its essential similarity to MSSC. Clearly, the homography model is able to replicate its strong performance ($11.45\%$) on this real-world dataset despite facing much stronger perspective effects. While all our single-view models turned in substantially better results than the baseline approach, it is also evident from the percentage errors that each single-view model has difficulties in dealing with real-world effects. The various multi-view schemes, especially the co-regularization approach, can further improve the performance.


\vspace{-0.9cm}
\subsubsection{Qualitative Study}

\begin{figure*}[h]
\begin{center}
\includegraphics[width=0.92\linewidth]{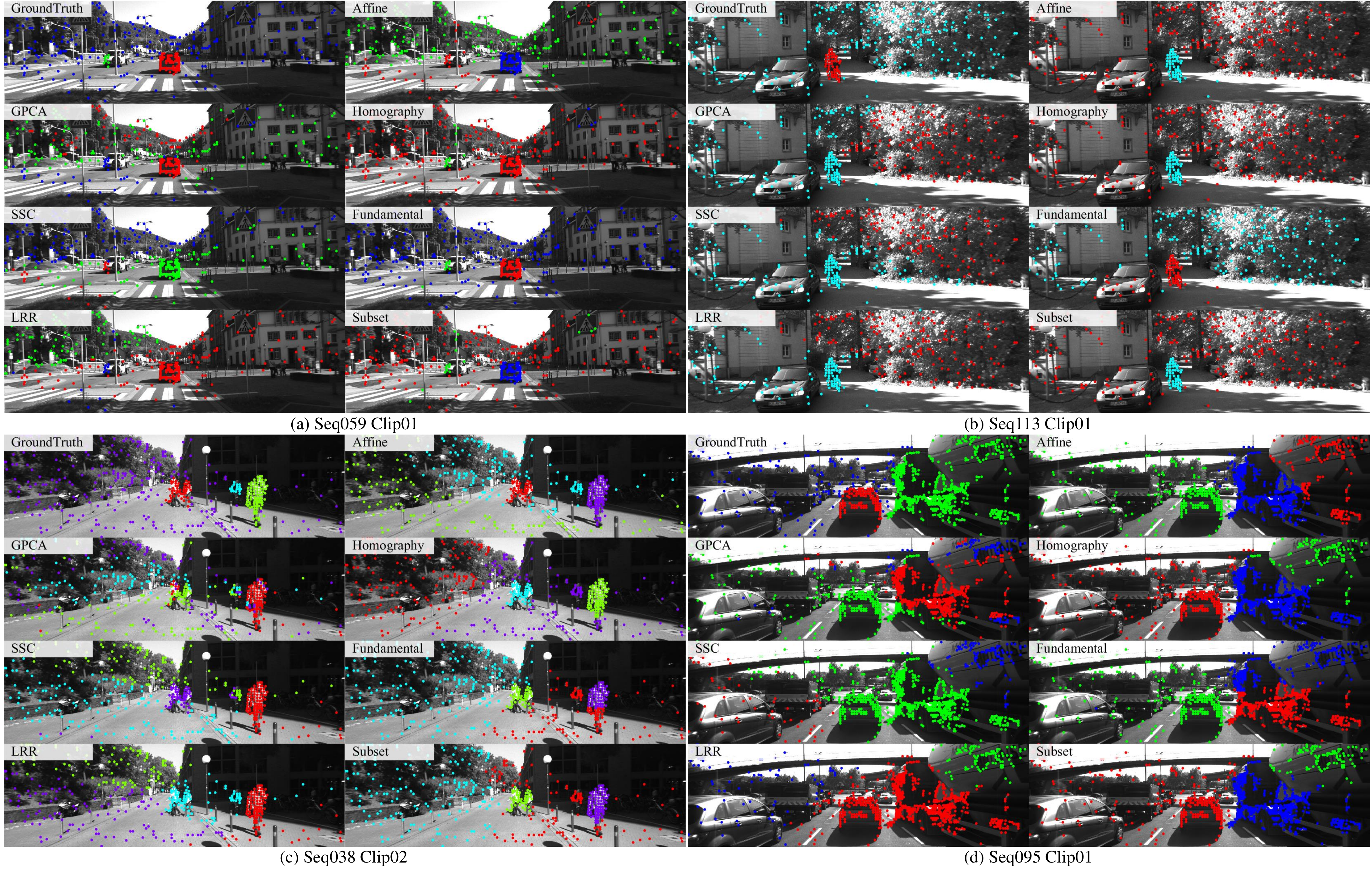}
\caption{Examples of motion segmentation on KT3DMoSeg sequences.}\label{fig:QualitativeKT3DMoSeg}
\vspace{-0.6cm}
\end{center}
\end{figure*}

We now present the motion segmentation results on some sequences from KT3DMoSeg in Fig.~\ref{fig:QualitativeKT3DMoSeg} to better understand how different geometric models complement each other, as well as to illustrate the challenges posed by this dataset. All these sequences involve strong perspective effects in the background but the foreground moving objects often have limited depth reliefs. Many background objects have non-compact shapes, and thus the slicing effect induced by the homography/affine model is less likely to relate all the background points together due to the lower connectivity. Therefore the background tends to split in the homography view, e.g. the traffic sign in Fig.~\ref{fig:QualitativeKT3DMoSeg} (a). While fundamental matrix is more likely to discover a seamless background in theory, it is plagued by a greater susceptibility to subspace overlap in practice. For instance, the scene in Fig.~\ref{fig:QualitativeKT3DMoSeg} (b) seems to be a classic scene to which the fundamental matrix is suited, and it seems here that even though a correct fundamental matrix for the background has been estimated (manifest by the blue cluster capturing both distant points as well as the tree nearby), the overlap between the foreground cyclist and the static car means that they are wrongly grouped together. In both (a) and (b), the fusion schemes manage to correct these errors. There are also some challenges that remain in this dataset. Clearly, when the motion of the foreground object (e.g. the person in the middle of Fig.~\ref{fig:QualitativeKT3DMoSeg} (c), indicated by blue points in GroundTruth) is small or intermittent compared to that of the camera, it can be difficult to detect. Coupled with the the large depth range in the background, the algorithm can be fooled to split the background instead of segmenting the foreground. Lastly, scenes like Fig.~\ref{fig:QualitativeKT3DMoSeg} (d)  still poses serious challenge. It is well known that the epipolar constraint allows a freedom to translate along the epipolar line. This allows an independent motion that is moving with respect to the background but consistent with the epipolar constraint to go undetected. In the figure, the car in front can be interpreted as a background object on the horizon, and thus the algorithm ends up splitting the big truck instead.

\subsection{Further Analysis}

\paragraph{Fusion Impact on Individual Views} As a result of the co-regularization, each of the geometric models has their views modified; we call these the F-view, H-view, and A-view. We now analyze the performance gain experienced by these views. In particular, we investigate the performance of motion segmentation with the spectral embedding of these views after co-regularization. This is equivalent to using just a single $\matr{U}=\matr{U}_v$ for k-means clustering in the last step of Algorithm~\ref{alg:Subset}. The classification error over all KT3DMoSeg sequences v.s. $\lambda$ and $\gamma$ are presented in Fig.~\ref{fig:KTMoSeg_PerfPerSeq} (b-c). We observe from this evaluation that while the F-view (purple line) does not necessarily produce the best result compared with the H-view without co-regularization, under certain range of $\lambda$ (corresponding to different coerciveness of the co-regularization), the F-view can be corrected so that its full potential is realized, producing the best of all results.

\vspace{-0.2cm}
\section{Conclusion}
In this paper, we have contributed to an understanding of the strengths and drawbacks of homography and fundamental matrices as a geometric model for motion segmentation, not only in the extant datasets such as Hopkins155, but also for real-world sequences in KT3DMoSeg. Not only do we account for the unexpected success of the homography approach when the affinities are accumulated to over all slicing planes, we also reveal its real limitation in real-world scenes. The geometrical exactness of the fundamental matrix approach is theoretically appealing; we show how its potential can be harnessed in a multi-view spectral clustering fusion scheme. Given kernels induced from multiple types of geometric models, we evaluate several techniques to synergistically fuse them. 
Finally, we carry out experiments on Hopkins155, Hopkins12  and MTPV62 and achieved state-of-the-art performances on all of them. In light of the demand for real-world motion segmentation, we further propose a new dataset, the KT3DMoSeg dataset, to reflect and investigate real challenges in motion segmentation in the wild.

{\small
\bibliographystyle{ieee}
\bibliography{cvpr18_1}
}

\end{document}